\documentclass[letterpaper,10pt,conference]{ieeeconf}

\usepackage{amsmath,amssymb,mathtools}
\usepackage{graphicx}
\usepackage[caption=false,font=footnotesize]{subfig}
\usepackage{booktabs,array,tabularx}
\usepackage[table]{xcolor}
\usepackage{placeins}
\usepackage{cite}

\definecolor{PSSBlue}{HTML}{0072B2}
\definecolor{PSSOrange}{HTML}{D55E00}
\definecolor{Ink}{HTML}{1D2733}

\newcommand{\pss}{\textsc{PSS}}
\newcommand{\R}{\mathbb{R}}
\newcommand{\E}{\mathbb{E}}

\makeatletter
\newcommand{\bottomtablecaption}[2]{%
  \begingroup
  \def\fnum@table{Table~\thetable}%
  \long\def\@makecaption##1##2{%
    \vskip\abovecaptionskip
    \parbox[t]{\hsize}{\footnotesize\noindent ##1.~~##2}}%
  \caption{#1}%
  \label{#2}%
  \endgroup
}
\makeatother

\title{\LARGE \bf Principal Steering Subspaces for Online Adaptation of\\
Frozen Generative Robot Policies}
\IEEEoverridecommandlockouts
\author{
Jialeng Ni$^{1,2}$,
Nathan Zhao$^{2,*}$,
and Kunpeng Song$^{2}$%
\thanks{\protect\raggedright $^{1}$University of Michigan, Ann Arbor, MI, USA.\protect\newline
{\ttfamily\footnotesize jialeng@umich.edu}\protect\endgraf}%
\thanks{\protect\raggedright $^{2}$XPENG Robotics, Santa Clara, CA, USA.\protect\newline
{\ttfamily\footnotesize \{jialeng.ni,nathan.zhao,kunpeng.song\}@xiaopeng.com}\protect\endgraf}%
\thanks{\protect\raggedright $^{*}$Corresponding author: Nathan Zhao
({\ttfamily\footnotesize zhaonat@gmail.com}).\protect\endgraf}%
}

\begin{document}
\maketitle
\thispagestyle{empty}
\pagestyle{empty}

\begin{abstract}
Generative robot policies provide expressive behavior priors, but updating a
large diffusion or flow-matching model through online interaction is costly.
Latent-space reinforcement learning avoids updating the pretrained generator by
controlling its initial sampling noise, yet high-dimensional noise can have
strongly anisotropic effects on decoded actions.  We introduce \emph{Principal
Steering Subspaces} (\pss{}), a forward-query interface that constructs a
fixed low-dimensional control basis from finite-difference decoder responses.
Soft Actor-Critic controls the leading response directions, while the orthogonal
complement is independently resampled from the Gaussian prior at each query.  On
three RoboMimic tasks with diffusion and flow-matching policies, response spectra
reveal substantial concentration.  Across five matched task--generator pairs,
the training curves indicate that \pss{} generally converges faster and exhibits
more stable late-training behavior than full-latent control, while achieving
stronger final performance overall.
Controlled Diffusion-Square ablations further show that leading-response
directions outperform random and least-responsive subspaces of equal dimension.
We further integrate \pss{} with
a frozen, closed-source 3B-parameter vision-language-action (VLA) policy in a
humanoid learning system with synchronous transition collection, reset-time
optimization, and latency-aware asynchronous deployment.  In an exploratory
screwdriver-placement evaluation, success is observed in 2/10 trials for the
frozen VLA policy and 6/10 after SAC+\pss{} adaptation.  These results support
decoder-response geometry as a practical basis for online adaptation of frozen
generative robot policies.
\end{abstract}

\section{Introduction}
\label{sec:intro}

Diffusion and flow-matching policies are powerful behavior priors for robot
manipulation: they can represent multimodal actions and generate temporally
coherent action chunks \cite{chi2023diffusionpolicy,black2024pi0}.  Yet a
pretrained policy inevitably encounters downstream objects, scenes, and task
preferences that differ from its training distribution.  Online reinforcement
learning (RL) can specialize the policy from interaction, but updating a large
generator on the robot is computationally demanding and risks degrading useful
pretrained behavior.

Diffusion Steering via Reinforcement Learning (DSRL) offers an appealing
alternative by freezing the generator and treating its initial sampling noise
as the action of an RL policy \cite{wagenmaker2025dsrl}.  The frozen model
becomes a nonlinear decoder from a latent RL action to a physical action chunk,
eliminating backpropagation through the generator.  This efficiency comes with a
control problem that has received less attention.  Full-latent SAC must explore
and fit value functions over all $D$ initial-noise coordinates, a dimension that
typically grows with the action chunk.  Those coordinates are also
geometrically misaligned with behavior: the decoder may be
highly sensitive to some combinations of noise and nearly invariant to others.
Isotropic exploration can therefore spend scarce robot interaction on latent
changes that produce little change in physical motion.

\begin{figure*}[t]
  \centering
  \includegraphics[width=\textwidth]{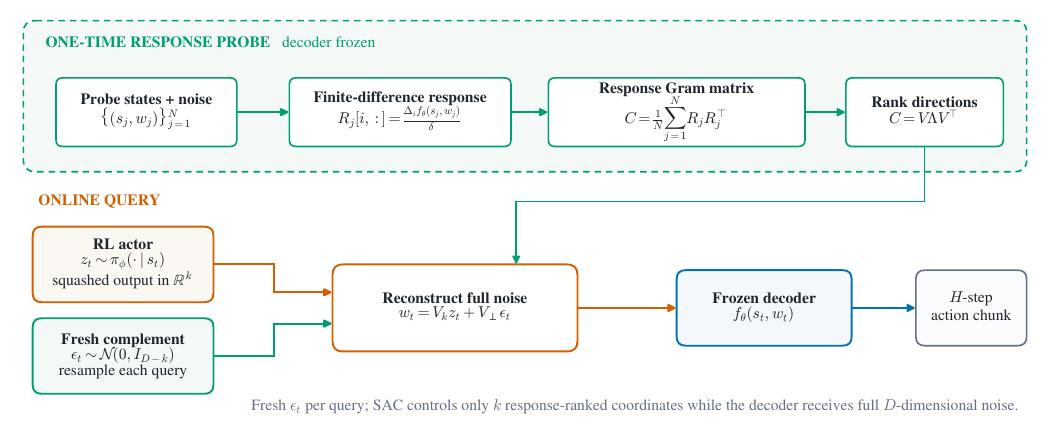}
  \caption{\textbf{Principal steering subspaces.}  A one-time finite-difference
  probe estimates a response Gram matrix and partitions its eigenbasis into
  controlled directions $V_k$ and complement $V_\perp$.  During online
  learning, SAC emits $z_t$, a fresh
  $\epsilon_t\sim\mathcal N(0,I_{D-k})$ fills the complement, and the frozen
  generator decodes $w_t=V_kz_t+V_\perp\epsilon_t$ into an action chunk.}
  \label{fig:overview}
\end{figure*}

We address this mismatch with \emph{Principal Steering Subspaces} (\pss{}),
illustrated in Fig.~\ref{fig:overview}.  Before online learning, finite
differences measure how initial-noise perturbations change decoded action chunks
over representative observations.  The leading eigenvectors of the resulting
response Gram matrix define a compact action $z\in\R^k$ for SAC.  At each query,
the orthogonal complement is resampled from the original Gaussian prior, so the
generator retains stochastic variation without requiring the actor and critics
to control every latent coordinate.  Rather than changing the frozen decoder or
the SAC update rule, \pss{} changes the coordinates exposed to RL: the actor
controls a fixed response-ranked subspace, while the remaining coordinates are
sampled independently from the generator prior.  The construction is independent
of the generator's training objective, requires only forward queries to
$f_\theta(s,w)$ with user-specified initial noise, and provides a response-energy
diagnostic for selecting $k$.

Hardware deployment introduces a complementary timing challenge.  Continuous
motion benefits from asynchronous inference, whereas sound RL transitions
require an unambiguous correspondence between an observation, a latent action,
and the resulting next observation.  We resolve this tension with synchronous
queries during data collection and latency-aware asynchronous queries during
deployment.  A state machine assigns terminal labels and schedules SAC updates
during scene reset while preserving transition semantics.

This paper makes three contributions:
\begin{itemize}
  \item We formulate \pss{}, a response-ranked latent action space that reduces
  SAC control from $D$ to $k$ dimensions while retaining prior stochasticity in
  the orthogonal complement, and relate its basis to a maximum-response
  variational objective.
  \item We evaluate the response geometry and online-learning behavior of
  frozen diffusion and flow-matching policies on three RoboMimic tasks.
  Controlled basis and dimension ablations distinguish the benefit of response
  ranking from dimensionality reduction alone.
  \item We demonstrate an end-to-end robot learning system with synchronous
  transition collection, reset-time optimization, and latency-aware
  asynchronous deployment, together with an exploratory humanoid adaptation
  study.
\end{itemize}

\section{Related Work}
\label{sec:related}

\paragraph{Generative policies and post-training}
Diffusion and flow-matching policies generate action chunks through iterative
denoising or ODE integration \cite{chi2023diffusionpolicy,lipman2023flowmatching,
black2024pi0}.  Post-training methods either optimize the generative process
itself with RL \cite{dppo2024,reinflow2025} or preserve the base model and learn
a smaller steering interface \cite{policydecorator2024,wagenmaker2025dsrl}.
DSRL is closest to our setting: it freezes the generator and treats the initial
noise as the RL action.  We likewise keep the generator frozen while changing
the geometry of the action presented to the actor and critics.

\paragraph{Latent steering for robot policies}
Latent interfaces for generative robot policies include learned bottlenecks and
RL tokens, perturbation-based diffusion steering, single-vector adaptation,
unified noise steering, and flow-reversal steering
\cite{zprl2026,rltoken2026,lpds2026,goldenticket2026,unisteer2026,frs2026}.
These methods differ in what is adapted and how the controllable coordinates are
obtained.  \pss{} addresses the coordinate-selection problem directly: it
constructs a fixed orthogonal basis from the decoded-action response of the
frozen decoder, then pairs that interface with an otherwise standard SAC
learner.

\paragraph{Low-dimensional control interfaces}
Latent-action and residual-policy methods reduce online search by learning or
reusing structure from offline data
\cite{spirl2020,plas2020,laser2021,johannink2019residualrl}.  Active-subspace
methods instead rank input directions by average local sensitivity
\cite{constantine2015active}.  \pss{} brings this response-based view to the
initial noise of a frozen action generator.  Unlike a learned action
autoencoder, the basis comes from finite differences of decoded action chunks;
unlike a deterministic bottleneck, the uncontrolled complement remains
stochastic under the generator prior.

\paragraph{Online robot learning systems}
SERL and HIL-SERL show that replay, high update-to-data ratios, and structured
human interaction can make real-robot actor--critic learning practical
\cite{luo2024serl,luo2025hilserl}.  Action chunking and temporal ensembling have
also proved effective for continuous execution of learned manipulation policies
\cite{zhao2023aloha}.  Our system connects these concerns to latent-space RL by
separating synchronous transition collection from asynchronous deployment, and
by scheduling updates inside an explicit reset/label state machine.  RoboMimic's
multi-human demonstrations and robosuite environments provide the simulation
testbed \cite{mandlekar2021robomimic,zhu2020robosuite}.

\section{Method}
\label{sec:method}

\subsection{Frozen Generative Policy as an RL Environment}

Let $s_t$ be the policy observation and let a frozen generator
$f_\theta(s_t,w_t)$ map initial noise $w_t\in\R^D$ to a flattened $H$-step
action chunk in $\R^M$, where $M=HA$ for action width $A$.  The formulation does
not require $D=M$.  A diffusion base is trained to predict denoising noise and
is sampled with DDIM \cite{song2021ddim}.  Our flow base uses the linear path
$x_\tau=(1-\tau)w+\tau a$ and target velocity $v=a-w$, then integrates the
learned velocity field with fixed-step Euler updates
\cite{rectifiedflow2022,lipman2023flowmatching}.  In both cases the generator is
deterministic conditional on $(s,w)$ during RL; only $w$ is controlled.

Following DSRL, SAC treats this noise as its action
\cite{haarnoja2018sac,wagenmaker2025dsrl}.  A squashed Gaussian actor samples
$w_t\sim\pi_\phi(\cdot\mid s_t)$, the frozen generator decodes a physical action
chunk, and replay stores $(s_t,w_t,r_t,s_{t+1},d_t)$.  For twin critics,
\begin{align}
y_t &= r_t+\gamma(1-d_t)\bigl[\min_i Q_{\bar\theta_i}(s_{t+1},w')
      \nonumber\\[-2pt]
    &\hspace{30mm}-\alpha\log\pi_\phi(w'\mid s_{t+1})\bigr],
      \label{eq:sac-target}\\
J_\pi&=\E_{s,w\sim\pi_\phi}
      [\alpha\log\pi_\phi(w\mid s)-\min_iQ_{\theta_i}(s,w)].
\end{align}
In simulation, $t$ indexes one policy query: up to $H$ low-level actions are
executed, their rewards are summed into $r_t$, and $\gamma$ is applied once at
the query boundary.  Reported environment steps count these executions across
the vectorized environments.  Direct control uses all $D$ coordinates.  For
RoboMimic Lift, Can, and Square, $H=4$, $A=7$, and $D=M=28$.

\subsection{Principal Steering Subspaces}

The central observation is that equal-size changes in different noise
directions need not produce equal changes in the decoded action.  Before RL, we
collect $N$ representative states by rolling out the frozen policy with i.i.d.
standard-Gaussian initial noise and independently sample one base noise
$w_j\sim\mathcal N(0,I_D)$ per state.  For coordinate $i$, a forward finite
difference estimates one row of the local decoded response
\begin{equation}
R_j[i,:]=\frac{f_\theta(s_j,w_j+\delta e_i)-f_\theta(s_j,w_j)}{\delta}.
\label{eq:fd}
\end{equation}
Thus, $R_j^\top\in\R^{M\times D}$ approximates the Jacobian of the decoded
action chunk with respect to initial noise, so that
$\Delta a\approx R_j^\top\Delta w$.  For RoboMimic, Eq.~\eqref{eq:fd} measures
the normalized, decoder-clipped chunk before environment unnormalization.  The
resulting positive-semidefinite response Gram matrix is
\begin{align}
C&=\frac{1}{N}\sum_{j=1}^{N}R_jR_j^\top,\nonumber\\[-2pt]
C&=V\operatorname{diag}(\lambda_1,\ldots,\lambda_D)V^\top,
\quad \lambda_1\geq\cdots\geq\lambda_D.
\label{eq:response}
\end{align}
Unlike a statistical covariance, $C$ is an uncentered average of squared local
responses.  Computing it requires $N(D+1)$ frozen-policy evaluations, which we
batch across states and perturbations.  We use $N=64$ and $\delta=0.05$ for the
RoboMimic spectra.  We compute one basis for each task and frozen checkpoint
before RL and keep it fixed throughout adaptation.

Let $V_k$ contain the first $k$ eigenvectors and $V_\perp$ the remainder.  The
SAC actor now outputs only $z_t\in\R^k$; every decode samples a fresh complement
$\epsilon_t\sim\mathcal N(0,I_{D-k})$ and reconstructs
\begin{equation}
w_t=V_kz_t+V_\perp\epsilon_t.
\label{eq:pss}
\end{equation}
For \pss{}, replay stores $(s_t,z_t,r_t,s_{t+1},d_t)$, and the SAC target and
actor loss above are evaluated in $z$-space; the independently sampled
complement is part of the transition stochasticity.
Thus the complement is environment stochasticity rather than a controllable RL
action.  The actor and critics operate on $z_t$, while the frozen generator
still receives a full $D$-dimensional noise tensor.  When $k=D$, the method
reduces to full-dimensional control in the response eigenbasis; with bounded
actor outputs this is a rotated box and should not be assumed identical to the
original axis-aligned control.

For any orthonormal $U\in\R^{D\times k}$ and
$\xi\sim\mathcal N(0,I_k)$, the probe-averaged squared linearized response is
\begin{equation}
\frac{1}{N}\sum_{j=1}^{N}\E_\xi
\!\left[\left\|R_j^\top U\xi\right\|_2^2\right]
=\operatorname{tr}(U^\top C U).
\label{eq:max-response}
\end{equation}
By the Ky Fan variational principle, $V_k$ maximizes this quantity and attains
$\sum_{i=1}^k\lambda_i$.  We therefore report
\begin{equation}
E(k)=\frac{\sum_{i=1}^k\lambda_i}{\sum_{i=1}^D\lambda_i}
\label{eq:energy}
\end{equation}
against the isotropic reference $k/D$.  This is a local response criterion, not
a reward-optimality guarantee: a low-energy direction can still matter for
contact or sparse reward.

\paragraph{Prior preservation and its limit}
If $z$ and $\epsilon$ are independent, with $z\sim\mathcal N(0,I_k)$ and
$\epsilon\sim\mathcal N(0,I_{D-k})$, the orthogonal reconstruction in
Eq.~\eqref{eq:pss} is exactly standard Gaussian.
The learned SAC actor is bounded and non-Gaussian, so only the complement has
this exact prior property in the deployed method.  We make no stronger claim
that every learned noise vector is in-distribution.

\section{Real-Robot Learning System}
\label{sec:system}

\begin{figure*}[t]
  \centering
  \subfloat[Synchronous training query\label{fig:system-sync}]{%
    \includegraphics[width=.485\textwidth]{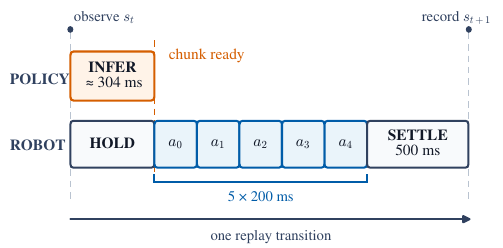}}%
  \hfill
  \subfloat[Latency-aware asynchronous deployment\label{fig:system-async}]{%
  \includegraphics[width=.485\textwidth]{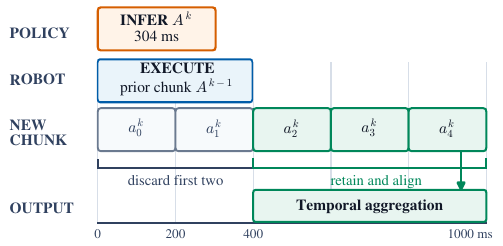}}
  \caption{\textbf{Query schedules for training and deployment.}
  (a) During online training, inference blocks execution so that
  $s_t$, the complete five-waypoint chunk, and $s_{t+1}$ define one unambiguous
  replay transition.  (b) During deployment, the next query runs while
  the current chunk executes.  The first two returned waypoints are no longer
  used under the measured query latency and are discarded; the aligned
  remainder is temporally aggregated with previous predictions before
  publication.}
  \label{fig:query-timing}
\end{figure*}

\begin{figure*}[t]
  \centering
  \includegraphics[width=.98\textwidth]{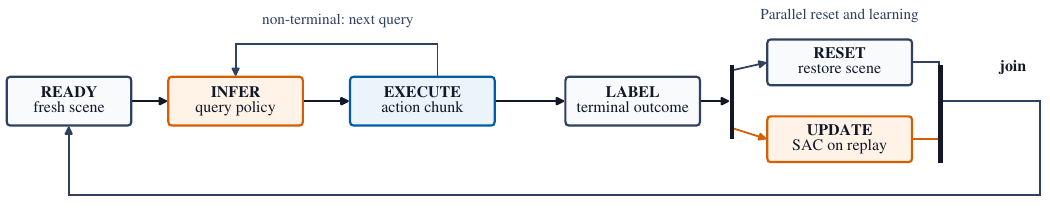}
  \caption{\textbf{Real-robot online RL interaction loop.}  The
  \textsc{infer}--\textsc{execute} cycle repeats until a terminal outcome is
  labeled.  Scene reset and SAC optimization then run concurrently, and a join
  barrier admits the next \textsc{ready} state only after both branches finish.}
  \label{fig:system-loop}
\end{figure*}

Deploying latent-space RL requires separating two concerns that are easy to
conflate: the temporal semantics of a training transition and the scheduling of
policy inference during continuous execution.  Figure~\ref{fig:query-timing}
contrasts the corresponding query schedules; Fig.~\ref{fig:system-loop} shows
how they fit into the complete interaction state machine.

\subsection{Synchronous Training, Asynchronous Deployment}

\paragraph{Training query}
During online RL, inference is synchronous.  At a fresh decision boundary the
actor samples $z_t$, \pss{} reconstructs $w_t$, and the frozen generator returns
an action chunk.  The robot remains stationary until this computation finishes,
then executes a fixed number of low-level actions before recording $s_{t+1}$.
This reduces motion-induced observation staleness during inference and makes
each replay entry correspond to one completed query/rollout block.  During
hardware training, $t$ indexes synchronous generator queries: one transition
spans inference, five 200-ms waypoints, and a subsequent 500-ms settle.

\paragraph{Deployment query}
Final inference need not block the robot.  While the current action chunk is
executed, an asynchronous worker queries the policy from the most recently
acquired observation.
The measured model latency is slightly above 300 ms, whereas each waypoint
spans 200 ms.  We therefore discard the first two points of every returned
chunk, align the remaining points to their intended execution times, and insert
them into a temporal-aggregation buffer.  For a given command time, predictions
$a_i$ are weighted relative to the newest $a_{\rm new}$ in proportion to
$\exp[5\,\operatorname{cos}(a_i,a_{\rm new})]$; fractional offsets are linearly
interpolated onto the 200-ms grid.  This coordinates successive predictions
while action chunking keeps the robot moving during inference.

\subsection{Interaction State Machine and Reset-Time Learning}

The robot loop uses explicit \textsc{reset}, \textsc{ready}, \textsc{infer},
\textsc{execute}, \textsc{label}, and \textsc{update} phases.  Inference and
execution repeat until an episode terminates, and success or failure is assigned
only at that boundary.  Updates run while the operator restores the scene, and a
join barrier prevents the next \textsc{ready} phase from starting with a
partially updated actor.
The state machine therefore overlaps computation with unavoidable reset work
without changing the semantic boundary of the collected transition.

\section{Experiments}
\label{sec:experiments}

We study four questions: (1) Is the response of a frozen generator concentrated
in a low-dimensional subspace? (2) Does controlling that subspace improve
online learning over full-latent control? (3) How do basis ranking and subspace
dimension affect the result? (4) Can the method operate in a real-robot learning
loop?

\paragraph{Evaluation protocol}
Each simulation condition uses one training seed.  A point on a learning curve
is the empirical success rate of one checkpoint over 200 stochastic evaluation
episodes; curves are unsmoothed.  One replay transition is an $H=4$ action
chunk; the horizontal axis counts its low-level actions across four training
environments.  Evaluations occur every 48,000 such steps for Lift and Can and
64,000 for Square.  Summaries average the final five evaluations at the run end,
except Flow-Lift, which uses the final five at or before $2\times10^6$ steps
within runs continuing to approximately $4\times10^6$.  This reduces
checkpoint-level evaluation noise but not variability across training seeds.

\subsection{RoboMimic Setup}

We use the low-dimensional multi-human RoboMimic datasets, each containing 300
demonstrations collected by six operators of varied proficiency in
robosuite/MuJoCo \cite{mandlekar2021robomimic,zhu2020robosuite}.  The released
HDF5 data are processed by the DPPO pipeline into min/max-normalized $[-1,1]$
arrays.  Observations contain end-effector pose, gripper state, and task-specific
object state; actions are operational-space pose increments and a gripper
command.  Table~\ref{tab:setup} summarizes the task-dependent data and model
dimensions.

\begin{table*}[t]
\centering
\footnotesize
\setlength{\tabcolsep}{5.5pt}
\begin{tabular}{lrrrrccc}
\toprule
Task & Transitions & Mean length & State dim. & Action dim. & Hidden MLP & Time dim. & Diff. train steps \\
\midrule
Lift   & 31,127 & 104 & 19 & 7 & $512\times3$ & 16 & 20 \\
Can    & 62,756 & 209 & 23 & 7 & $512\times3$ & 16 & 20 \\
Square & 80,731 & 269 & 23 & 7 & $1024\times3$ + cond. & 32 & 100 \\
\bottomrule
\end{tabular}
\vspace{1mm}

\begin{tabularx}{.98\textwidth}{@{}>{\raggedright\arraybackslash}p{.15\textwidth}>{\raggedright\arraybackslash}p{.31\textwidth}>{\raggedright\arraybackslash}p{.16\textwidth}>{\raggedright\arraybackslash}X@{}}
\toprule
Generator setting & Value & Online setting & Value \\
\midrule
Flow time sampling & $t=(0.999-u)/0.999$, $u\sim\operatorname{Beta}(1.5,1)$
& SAC learning rates & $3\times10^{-4}$ (actor, critics, temperature) \\
Flow optimization & AdamW, $10^{-4}$ learning rate, $10^{-6}$ weight decay,
batch size 256 & Batch size; target $\tau$ & 256; 0.005 \\
Flow schedule & 3000 epochs, cosine decay after 100 warmup steps, EMA 0.995
& Discount; updates/transition & 0.99 (Square: 0.999); 30 (Lift), 20 otherwise \\
Inference & Eight function evaluations for both generator families
& \pss{} probes; primary $k$ & $N=64$, $\delta=0.05$; $k=8$ \\
\bottomrule
\end{tabularx}
\bottomtablecaption{\textbf{Simulation datasets and configurations.}
All tasks use 300 multi-human demonstrations and action-chunk horizon $H=4$.
``cond.'' denotes the Square observation-conditioning MLP with widths
$[512,64]$; time dim. is the time-embedding width.}{tab:setup}
\end{table*}

The diffusion bases are released DDPM $\epsilon$-prediction checkpoints sampled
with deterministic DDIM.  The flow bases use the same DiffusionMLP backbone,
the linear conditional-flow path described in Sec.~\ref{sec:method}, and
fixed-step Euler integration.  We train the flow models with the configuration
in Table~\ref{tab:setup}, selecting the Lift and Square
checkpoints at epoch 2500 and the Can checkpoint at epoch 3000 by open-loop
action MSE.  Both generator families clamp decoded actions to $[-1,1]$.
The flow-time expression in Table~\ref{tab:setup} is used without additional
clipping: its support extends to approximately $-0.001$, and 0.15\% of samples
fall in this short extrapolation below zero.

SAC uses twin critics and automatic entropy tuning.  Table~\ref{tab:setup}
lists the settings shared by each full-latent/\pss{} pair.  For Flow-Square, the
\pss{} and full-latent runs use controllable-noise bounds of $[-1,1]$ and
$[-1.5,1.5]$, respectively.

\subsection{Decoder Response Geometry}

\begin{figure*}[t]
  \centering
  \subfloat[Diffusion policies\label{fig:pss-analysis-diffusion}]{%
    \includegraphics[width=.485\textwidth]{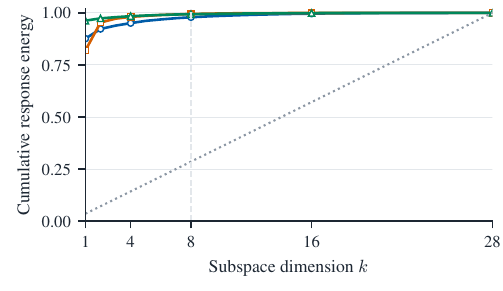}}%
  \hfill
  \subfloat[Flow-matching policies\label{fig:pss-analysis-flow}]{%
    \includegraphics[width=.485\textwidth]{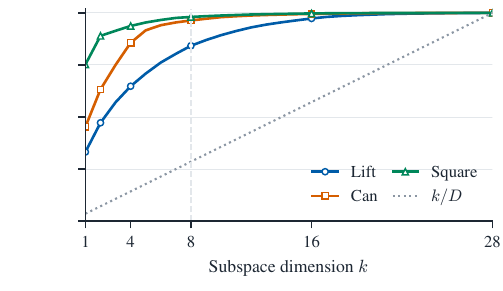}}
  \caption{\textbf{Decoder-response spectra.}
  (a) Cumulative finite-difference response energy for the diffusion
  policy on Lift, Can, and Square.  (b) The corresponding spectra for
  the flow-matching policy.  In both panels, the dotted diagonal is the
  isotropic reference $k/D$, and the vertical line marks the common operating
  dimension $k=8$.}
  \label{fig:pss-analysis}
\end{figure*}

Figure~\ref{fig:pss-analysis} shows that response energy is substantially more
concentrated than the isotropic reference.  At $k=8$, the diffusion policies
retain at least 97.9\% of measured response, while the flow policies retain
84.2--98.1\%; all exceed the isotropic value of 28.6\%.  Flow-Lift is the least
concentrated case, indicating that the degree of compression depends on both
task and generator.  These spectra motivate a compact control interface, but
they do not establish reward relevance: a weak local response may still be
important near contact.  The ablations in Sec.~\ref{sec:basis-ablation}
therefore test basis choice and dimension through online learning.

\subsection{Full-Latent Control versus PSS}

The full-latent condition implements DSRL with SAC acting directly on all
$D=28$ initial-noise coordinates.  The proposed condition keeps the same SAC
configuration but exposes only the leading eight response coordinates and
resamples the complement at every decode.

\begin{figure*}[t]
  \centering
  \includegraphics[width=.42\textwidth]{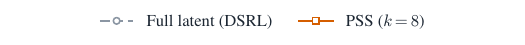}\\[-1mm]
  \subfloat[Lift\label{fig:diff-lift}]{%
    \includegraphics[width=.318\textwidth]{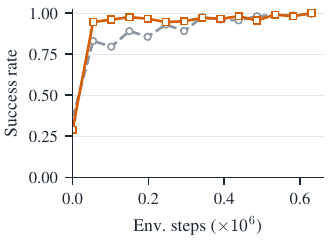}}%
  \hfill
  \subfloat[Can\label{fig:diff-can}]{%
    \includegraphics[width=.318\textwidth]{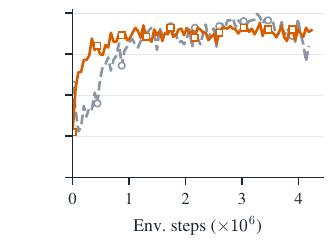}}%
  \hfill
  \subfloat[Square\label{fig:diff-square}]{%
    \includegraphics[width=.318\textwidth]{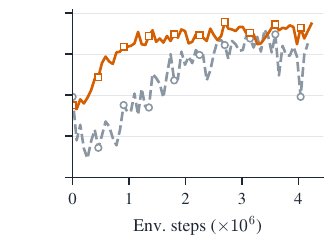}}
  \caption{\textbf{Online SAC with frozen diffusion policies.}
  (a) Lift, (b) Can, and (c) Square compare
  full-latent DSRL with \pss{} at $k=8$.  Each point is a 200-episode
  evaluation of a checkpoint from one training seed; lines connect unsmoothed
  evaluations.}
  \label{fig:diff-results}
\end{figure*}

\begin{figure*}[t]
  \centering
  \includegraphics[width=.42\textwidth]{figures/fig_robomimic_legend.pdf}\\[-1mm]
  \subfloat[Lift\label{fig:flow-lift}]{%
    \includegraphics[width=.318\textwidth]{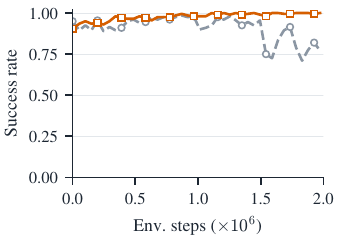}}%
  \hfill
  \subfloat[Can\label{fig:flow-can}]{%
    \includegraphics[width=.318\textwidth]{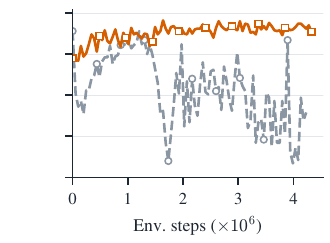}}%
  \hfill
  \subfloat[Square\label{fig:flow-square}]{%
  \includegraphics[width=.318\textwidth]{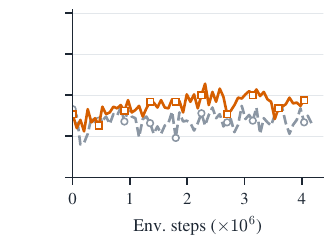}}
  \caption{\textbf{Online SAC with frozen flow-matching policies.}
  (a) Lift uses a $2\times10^6$-step reporting horizon within longer runs.
  Each point is a 200-episode evaluation of a checkpoint from one training seed;
  lines connect unsmoothed evaluations.}
  \label{fig:flow-results}
\end{figure*}

\begin{table}[t]
\centering
\footnotesize
\setlength{\tabcolsep}{4.5pt}
\begin{tabular}{llrrr}
\toprule
Generator & Task & Full latent & \pss{} & $\Delta$ (p.p.) \\
\midrule
Diffusion & Lift   & 0.981 & 0.981 & $\phantom{+}0.0$ \\
Diffusion & Can    & 0.831 & 0.884 & $+5.3$ \\
Diffusion & Square & 0.716 & 0.878 & $+16.2$ \\
Flow & Lift        & 0.773 & 0.999 & $+22.6$ \\
Flow & Can         & 0.302 & 0.914 & $+61.2$ \\
Flow & Square      & 0.362 & 0.459 & $+9.7$ \\
\bottomrule
\end{tabular}
\bottomtablecaption{\textbf{Final-window simulation results.}  Entries are mean
success rates over the final five evaluations at the reporting horizons defined
above; $\Delta$ is \pss{} minus full latent in percentage points.}{tab:main-results}
\end{table}

Table~\ref{tab:main-results} consolidates the endpoint metric, while
Figs.~\ref{fig:diff-results} and \ref{fig:flow-results} show the learning
dynamics that produce it.  We define the stable 80\% criterion as the third
checkpoint in the first sequence of three consecutive evaluations at or above
0.8.  Late-training variability is the population standard deviation over the
final 20\% of evaluated checkpoints; post-threshold regressions count later
evaluations below 0.8 after first reaching the threshold.  Across the five matched
task--generator pairs, \pss{} meets the stable criterion earlier in all four
non-ceiling cases and at the same checkpoint in the ceiling case.  It has a
higher final-window mean in four matched pairs and matches full-latent control on
Diffusion-Lift, with lower within-run late-training variability in four pairs and fewer
post-threshold regressions in all five.  The largest endpoint difference is on
Flow-Can, where both methods initially improve, but the full-latent run degrades
late while \pss{} remains stable.  This behavior is
consistent with a lower-dimensional critic being easier to fit, although the
present experiment does not isolate critic approximation error as the causal
mechanism.

\subsection{Basis and Dimension Ablations on Square}
\label{sec:basis-ablation}

Response concentration alone does not show that its leading directions are
better RL actions than another subspace of the same size.  We therefore compare
three $k=8$ bases on Diffusion-Square in a separate ablation campaign.  Its
response matrix and training rollout differ from the primary Diffusion-Square
\pss{} run in Table~\ref{tab:main-results}, explaining the respective 0.838 and
0.878 final-window means.  Within the campaign, all conditions share the frozen
checkpoint, response matrix, normalization statistics, SAC configuration,
training seed, and evaluation protocol.  Top-$k$ uses the leading response
eigenvectors, random-$k$ a Haar-distributed orthonormal basis, and least-$k$ the
smallest-response eigenvectors; all resample the same Gaussian complement.

We separately vary the controlled dimension over
$k\in\{2,4,8,16,28\}$ while always selecting the leading response directions.
For $k<28$, the complement is resampled as above; for $k=D=28$, SAC controls
every response-basis coordinate and no stochastic complement remains.

\begin{figure*}[t]
  \centering
  \subfloat[Basis selection at $k=8$\label{fig:basis-selection}]{%
    \includegraphics[width=.485\textwidth]{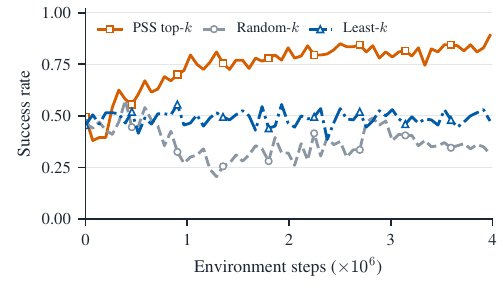}}%
  \hfill
  \subfloat[Controlled dimension\label{fig:dimension-selection}]{%
    \includegraphics[width=.485\textwidth]{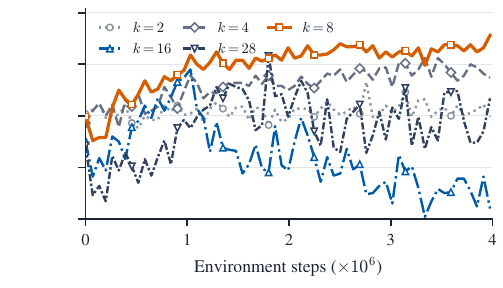}}
  \caption{\textbf{Diffusion-Square design ablations.}
  (a) The controlled dimension is fixed while the response-basis
  ranking is changed.  (b) The leading response basis is retained while
  the controlled dimension is varied.  The $k=8$ trace is shared by both panels.
  All curves show unsmoothed 200-episode evaluations from one matched seed
  through $4\times10^6$ environment steps.}
  \label{fig:basis-ablation}
\end{figure*}

\begin{table}[t]
\centering
\footnotesize
\setlength{\tabcolsep}{5pt}
\begin{tabular}{llr}
\toprule
Study & Setting & Final-five mean \\
\midrule
Basis & Top-$8$ & 0.838 \\
Basis & Random-$8$ & 0.346 \\
Basis & Least-$8$ & 0.498 \\
\addlinespace
Dimension & $k=2$ & 0.519 \\
Dimension & $k=4$ & 0.714 \\
Dimension & $k=8$ & 0.838 \\
Dimension & $k=16$ & 0.129 \\
Dimension & $k=28$ & 0.445 \\
\bottomrule
\end{tabular}
\bottomtablecaption{\textbf{Final-window ablation summary.}  Each entry averages
the final five evaluations at or before $4\times10^6$ steps in
Fig.~\ref{fig:basis-ablation}; the top-$8$ run is common to both
studies.}{tab:ablations}
\end{table}

At fixed dimension, the leading-response basis clearly outperforms both
controls (Fig.~\ref{fig:basis-selection} and Table~\ref{tab:ablations}).  Since
action dimension, complement sampling, and training budget are unchanged, the
separation cannot be explained by dimensionality reduction alone.  Least-$8$
does exceed Random-$8$, however, so the result supports selecting the leading
subspace rather than a monotonic relationship between response rank and return
for every possible basis.

The dimension sweep is similarly non-monotonic.  The $k=8$ run has the strongest
final window, whereas $k=2$ and $k=4$ omit more responsive directions and
$k=16$ and $k=28$ have lower final-window means.  In particular, $k=16$ reaches
competitive success earlier and then declines, a behavior hidden by reporting
only its peak.
These results support $k=8$ for this checkpoint and SAC configuration, but do
not imply a task-independent optimum.

\begin{figure*}[t]
  \centering
  \includegraphics[width=.98\textwidth]{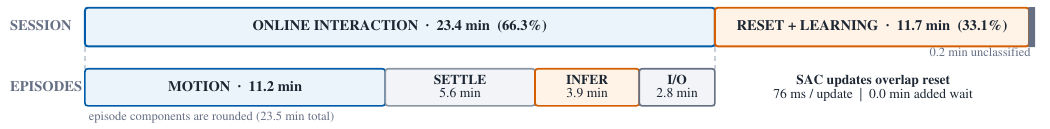}
  \caption{\textbf{Wall-clock profile of one 141-episode robot-training
  session.}  The upper bar partitions total session time into online interaction
  and the concurrent reset-and-learning window; the lower bar decomposes
  cumulative online-interaction time across the 141 episodes into motion,
  settling, inference, and environment/I/O work.  Durations are measured from
  runtime logs and rounded, so the displayed components differ slightly in their
  totals.}
  \label{fig:real-timing}
\end{figure*}

\begin{figure*}[t]
  \centering
  \subfloat[Frozen-VLA failure\label{fig:real-failure}]{%
    \includegraphics[width=.94\textwidth]{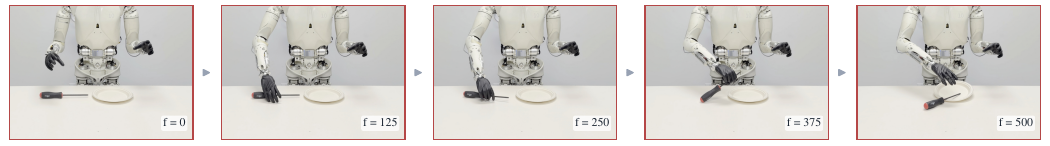}}\\[1.5mm]
  \subfloat[SAC+\pss{} success\label{fig:real-success}]{%
    \includegraphics[width=.94\textwidth]{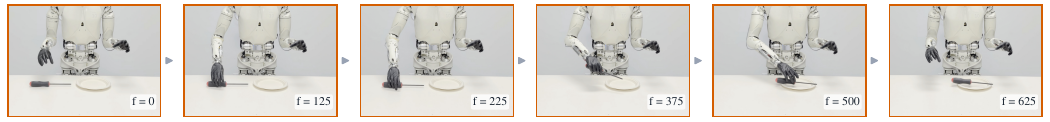}}
  \caption{\textbf{Representative screwdriver-placement trajectories.}
  (a) The frozen VLA selects an ineffective approach/grasp and does not
  complete placement.  (b) After online adaptation, SAC+\pss{} steers
  the frozen decoder to grasp and place the screwdriver successfully.  These
  sequences illustrate behavior; aggregate outcomes are reported in
  Table~\ref{tab:hardware}.}
  \label{fig:real-trajectories}
\end{figure*}

\subsection{Real-Robot Deployment}

The hardware study uses a frozen, closed-source 3B-parameter flow-matching
vision-language-action (VLA) policy on a humanoid robot for the task \emph{put
the screwdriver in the plate}.
The base policy often approaches or grasps the thin shaft instead of the handle,
providing a concrete affordance error for online correction.  We compare the
frozen policy alone with the same frozen VLA under the learned SAC+\pss{}
controller over ten evaluation trials per condition.  This hardware study
evaluates the feasibility of the complete SAC+\pss{} system; it does not include
a matched full-latent SAC baseline.  Consequently, comparisons between \pss{}
and full-latent control are restricted to simulation.
Figure~\ref{fig:real-timing} profiles one 141-episode robot-training session.
During hardware training, each synchronous query executes five 200-ms waypoints
followed by a 500-ms settle (Sec.~\ref{sec:system}).

\begin{table}[h]
\centering
\footnotesize
\setlength{\tabcolsep}{5pt}
\begin{tabular}{lrr}
\toprule
Policy & Successes/trials & Rate [95\% CI] \\
\midrule
Frozen VLA & 2/10 & 0.20 [0.057, 0.510] \\
SAC+\pss{} & 6/10 & 0.60 [0.313, 0.832] \\
\bottomrule
\end{tabular}
\bottomtablecaption{\textbf{Exploratory real-robot outcomes.}  Intervals are
two-sided 95\% Wilson intervals for evaluation-trial success.  The small study
is reported descriptively; it is not a statistically powered benchmark.}{tab:hardware}
\end{table}

The observed success rate is higher after adaptation
(Table~\ref{tab:hardware}), and the successful sequence in
Fig.~\ref{fig:real-trajectories} is consistent with correction of the grasp
selection error.  The intervals remain wide and overlap, so the result should be
read as evidence of feasibility rather than a precise estimate of improvement.

The timing profile in Fig.~\ref{fig:real-timing} shows that approximately two
thirds of the session is spent collecting interaction and one third restoring
the scene while learning proceeds.  Motion is the largest episode-internal
component; inference and settling account for most of the remainder.  SAC
updates averaged 76 ms and completed within the manual-reset interval, with no
measurable added wait at the logging resolution.

\paragraph{Limitations}
The simulation study uses one training seed per condition, and the basis and
dimension ablations cover only Diffusion-Square.  The response Gram matrix is a
local average over a fixed probe set and remains unchanged during learning.
The hardware study covers ten trials per condition on one task and lacks an
overlap-disabled timing control.  These limits preclude population-level or
causal timing claims.

\FloatBarrier

\section{Conclusion}
We introduced \pss{}, an RL interface derived from the response geometry of a
frozen generator.  Simulation shows concentrated sensitivity, a benefit from
response ranking beyond dimension reduction, and a non-monotonic dimension
tradeoff.  \pss{} also supports synchronous robot-training transitions,
reset-time updates, and asynchronous deployment.  One-seed simulation and the
small hardware study motivate multi-seed evaluation, broader robot experiments,
and state-adaptive response bases.

\bibliographystyle{IEEEtran}
\bibliography{references}

@article{chi2023diffusionpolicy,
  title   = {Diffusion Policy: Visuomotor Policy Learning via Action Diffusion},
  author  = {Chi, Cheng and Xu, Zhenjia and Feng, Siyuan and Cousineau, Eric and Du, Yilun and Burchfiel, Benjamin and Tedrake, Russ and Song, Shuran},
  journal = {The International Journal of Robotics Research},
  year    = {2025},
  volume  = {44},
  number  = {10--11},
  pages   = {1684--1704},
  doi     = {10.1177/02783649241273668},
}

@inproceedings{song2021ddim,
  title     = {Denoising Diffusion Implicit Models},
  author    = {Song, Jiaming and Meng, Chenlin and Ermon, Stefano},
  booktitle = {International Conference on Learning Representations},
  year      = {2021},
  eprint    = {2010.02502},
  archivePrefix = {arXiv},
}

@inproceedings{black2024pi0,
  title     = {{$\pi_0$}: A Vision-Language-Action Flow Model for General Robot Control},
  author    = {Black, Kevin and Brown, Noah and Driess, Danny and Esmail, Adnan and Equi, Michael and Finn, Chelsea and Fusai, Niccolo and Groom, Lachy and Hausman, Karol and Ichter, Brian and Jakubczak, Szymon and Jones, Tim and Ke, Liyiming and Levine, Sergey and Li-Bell, Adrian and Mothukuri, Mohith and Nair, Suraj and Pertsch, Karl and Shi, Lucy Xiaoyang and Tanner, James and Vuong, Quan and Walling, Anna and Wang, Haohuan and Zhilinsky, Ury},
  booktitle = {Robotics: Science and Systems},
  year      = {2025},
  eprint    = {2410.24164},
  archivePrefix = {arXiv},
}

@inproceedings{haarnoja2018sac,
  title     = {Soft Actor-Critic: Off-Policy Maximum Entropy Deep Reinforcement Learning with a Stochastic Actor},
  author    = {Haarnoja, Tuomas and Zhou, Aurick and Abbeel, Pieter and Levine, Sergey},
  booktitle = {Proceedings of the 35th International Conference on Machine Learning},
  year      = {2018},
  volume    = {80},
  series    = {Proceedings of Machine Learning Research},
  pages     = {1861--1870},
  publisher = {PMLR},
}

@inproceedings{luo2024serl,
  title     = {{SERL}: A Software Suite for Sample-Efficient Robotic Reinforcement Learning},
  author    = {Luo, Jianlan and Hu, Zheyuan and Xu, Charles and Tan, You Liang and Berg, Jacob and Sharma, Archit and Schaal, Stefan and Finn, Chelsea and Gupta, Abhishek and Levine, Sergey},
  booktitle = {IEEE International Conference on Robotics and Automation},
  year      = {2024},
  doi       = {10.1109/ICRA57147.2024.10610040},
  eprint    = {2401.16013},
  archivePrefix = {arXiv},
}

@article{luo2025hilserl,
  title   = {Precise and Dexterous Robotic Manipulation via Human-in-the-Loop Reinforcement Learning},
  author  = {Luo, Jianlan and Xu, Charles and Wu, Jeffrey and Levine, Sergey},
  journal = {Science Robotics},
  year    = {2025},
  volume  = {10},
  number  = {105},
  pages   = {eads5033},
  doi     = {10.1126/scirobotics.ads5033},
  eprint  = {2410.21845},
  archivePrefix = {arXiv},
}

@inproceedings{dppo2024,
  title     = {Diffusion Policy Policy Optimization},
  author    = {Ren, Allen Z. and Lidard, Justin and Ankile, Lars L. and Simeonov, Anthony and Agrawal, Pulkit and Majumdar, Anirudha and Burchfiel, Benjamin and Dai, Hongkai and Simchowitz, Max},
  booktitle = {International Conference on Learning Representations},
  year      = {2025},
  eprint    = {2409.00588},
  archivePrefix = {arXiv},
}

@inproceedings{reinflow2025,
  title     = {{ReinFlow}: Fine-tuning Flow Matching Policy with Online Reinforcement Learning},
  author    = {Zhang, Tonghe and Yu, Chao and Su, Sichang and Wang, Yu},
  booktitle = {Advances in Neural Information Processing Systems},
  year      = {2025},
  volume    = {38},
  eprint    = {2505.22094},
  archivePrefix = {arXiv},
}

@inproceedings{policydecorator2024,
  title     = {Policy Decorator: Model-Agnostic Online Refinement for Large Policy Model},
  author    = {Yuan, Xiu and Mu, Tongzhou and Tao, Stone and Fang, Yunhao and Zhang, Mengke and Su, Hao},
  booktitle = {International Conference on Learning Representations},
  year      = {2025},
  eprint    = {2412.13630},
  archivePrefix = {arXiv},
}

@inproceedings{wagenmaker2025dsrl,
  title     = {Steering Your Diffusion Policy with Latent Space Reinforcement Learning},
  author    = {Wagenmaker, Andrew and Zhang, Yunchu and Nakamoto, Mitsuhiko and Park, Seohong and Yagoub, Waleed and Nagabandi, Anusha and Gupta, Abhishek and Levine, Sergey},
  booktitle = {Proceedings of the 9th Conference on Robot Learning},
  year      = {2025},
  volume    = {305},
  series    = {Proceedings of Machine Learning Research},
  pages     = {258--282},
  publisher = {PMLR},
  eprint    = {2506.15799},
  archivePrefix = {arXiv},
}

@article{zprl2026,
  title   = {Beyond Action Residuals: Real-World Robot Policy Steering via Bottleneck Latent Reinforcement Learning},
  author  = {Yu, Dongjie and Lei, Kun and Jiang, Zhennan and Pan, Jia and Xu, Huazhe},
  journal = {arXiv preprint arXiv:2605.19919},
  year    = {2026},
  eprint  = {2605.19919},
  archivePrefix = {arXiv},
}

@article{rltoken2026,
  title   = {{RL Token}: Bootstrapping Online RL with Vision-Language-Action Models},
  author  = {Xu, Charles and Springenberg, Jost Tobias and Equi, Michael and Amin, Ali and Esmail, Adnan and Levine, Sergey and Ke, Liyiming},
  journal = {arXiv preprint arXiv:2604.23073},
  year    = {2026},
  eprint  = {2604.23073},
  archivePrefix = {arXiv},
}

@article{lpds2026,
  title   = {Lagrangian Perturbation Diffusion Steering: Latent Reinforcement Learning for Generative Policies},
  author  = {Simsir, Hikmet and Oguz, Ozgur S.},
  journal = {arXiv preprint arXiv:2606.01151},
  year    = {2026},
  eprint  = {2606.01151},
  archivePrefix = {arXiv},
}

@article{goldenticket2026,
  title   = {You've Got a Golden Ticket: Improving Generative Robot Policies With A Single Noise Vector},
  author  = {Patil, Omkar and Biza, Ondrej and Weng, Thomas and Schmeckpeper, Karl and Thomason, Wil and Zhang, Xiaohan and Sivakumar, Kausik and Walters, Robin and Gopalan, Nakul and Castro, Sebastian and Hart, Stephen and Rosen, Eric},
  journal = {arXiv preprint arXiv:2603.15757},
  year    = {2026},
  eprint  = {2603.15757},
  archivePrefix = {arXiv},
}

@article{unisteer2026,
  title   = {{UniSteer}: Unified Noise Steering for Efficient Human-Guided VLA Adaptation},
  author  = {Lu, Junjie and Qin, Xinyao and Jiang, Yuhua and Wang, Kaixin and Zhang, Chuheng and Liang, Bin and Yang, Jun and Xu, Min and Zhao, Li},
  journal = {arXiv preprint arXiv:2605.10821},
  year    = {2026},
  eprint  = {2605.10821},
  archivePrefix = {arXiv},
}

@article{frs2026,
  title   = {Improving Robotic Generalist Policies via Flow Reversal Steering},
  author  = {Tang, Andy and Chen, William and Wagenmaker, Andrew and Finn, Chelsea and Levine, Sergey},
  journal = {arXiv preprint arXiv:2606.13675},
  year    = {2026},
  eprint  = {2606.13675},
  archivePrefix = {arXiv},
}

@inproceedings{zhao2023aloha,
  title     = {Learning Fine-Grained Bimanual Manipulation with Low-Cost Hardware},
  author    = {Zhao, Tony Z. and Kumar, Vikash and Levine, Sergey and Finn, Chelsea},
  booktitle = {Proceedings of Robotics: Science and Systems},
  year      = {2023},
  doi       = {10.15607/RSS.2023.XIX.016},
  eprint    = {2304.13705},
  archivePrefix = {arXiv},
}

@book{constantine2015active,
  title     = {Active Subspaces: Emerging Ideas for Dimension Reduction in Parameter Studies},
  author    = {Constantine, Paul G.},
  publisher = {Society for Industrial and Applied Mathematics},
  address   = {Philadelphia, PA},
  year      = {2015},
  series    = {SIAM Spotlights},
  volume    = {2},
  isbn      = {978-1-61197-385-3},
  doi       = {10.1137/1.9781611973860},
}

@inproceedings{plas2020,
  title     = {{PLAS}: Latent Action Space for Offline Reinforcement Learning},
  author    = {Zhou, Wenxuan and Bajracharya, Sujay and Held, David},
  booktitle = {Proceedings of the 2020 Conference on Robot Learning},
  year      = {2021},
  volume    = {155},
  series    = {Proceedings of Machine Learning Research},
  pages     = {1719--1735},
  publisher = {PMLR},
}

@inproceedings{laser2021,
  title     = {{LASER}: Learning a Latent Action Space for Efficient Reinforcement Learning},
  author    = {Allshire, Arthur and Mart{\'i}n-Mart{\'i}n, Roberto and Lin, Charles and Manuel, Shawn and Savarese, Silvio and Garg, Animesh},
  booktitle = {IEEE International Conference on Robotics and Automation},
  year      = {2021},
  eprint    = {2103.15793},
  archivePrefix = {arXiv},
}

@inproceedings{spirl2020,
  title     = {Accelerating Reinforcement Learning with Learned Skill Priors},
  author    = {Pertsch, Karl and Lee, Youngwoon and Lim, Joseph J.},
  booktitle = {Proceedings of the 2020 Conference on Robot Learning},
  year      = {2021},
  volume    = {155},
  series    = {Proceedings of Machine Learning Research},
  pages     = {188--204},
  publisher = {PMLR},
}

@inproceedings{johannink2019residualrl,
  title     = {Residual Reinforcement Learning for Robot Control},
  author    = {Johannink, Tobias and Bahl, Shikhar and Nair, Ashvin and Luo, Jianlan and Kumar, Avinash and Loskyll, Matthias and Ojea, Juan Aparicio and Solowjow, Eugen and Levine, Sergey},
  booktitle = {IEEE International Conference on Robotics and Automation},
  year      = {2019},
  pages     = {6023--6029},
  doi       = {10.1109/ICRA.2019.8794127},
  eprint    = {1812.03201},
  archivePrefix = {arXiv},
}

@inproceedings{rectifiedflow2022,
  title     = {Flow Straight and Fast: Learning to Generate and Transfer Data with Rectified Flow},
  author    = {Liu, Xingchao and Gong, Chengyue and Liu, Qiang},
  booktitle = {International Conference on Learning Representations},
  year      = {2023},
  eprint    = {2209.03003},
  archivePrefix = {arXiv},
}

@inproceedings{lipman2023flowmatching,
  title     = {Flow Matching for Generative Modeling},
  author    = {Lipman, Yaron and Chen, Ricky T. Q. and Ben-Hamu, Heli and Nickel, Maximilian and Le, Matthew},
  booktitle = {International Conference on Learning Representations},
  year      = {2023},
  eprint    = {2210.02747},
  archivePrefix = {arXiv},
}

@inproceedings{mandlekar2021robomimic,
  title     = {What Matters in Learning from Offline Human Demonstrations for Robot Manipulation},
  author    = {Mandlekar, Ajay and Xu, Danfei and Wong, Josiah and Nasiriany, Soroush and Wang, Chen and Kulkarni, Rohun and Fei-Fei, Li and Savarese, Silvio and Zhu, Yuke and Mart{\'i}n-Mart{\'i}n, Roberto},
  booktitle = {Proceedings of the 5th Conference on Robot Learning},
  year      = {2021},
}

@article{zhu2020robosuite,
  title   = {robosuite: A Modular Simulation Framework and Benchmark for Robot Learning},
  author  = {Zhu, Yuke and Wong, Josiah and Mandlekar, Ajay and Mart{\'i}n-Mart{\'i}n, Roberto and Joshi, Abhishek and Lin, Kevin and Maddukuri, Abhiram and Nasiriany, Soroush and Zhu, Yifeng},
  journal = {arXiv preprint arXiv:2009.12293},
  year    = {2020},
}

\end{document}